\documentclass[conference]{IEEEtran}
\IEEEoverridecommandlockouts
\usepackage{cite}
\usepackage{amsmath,amssymb,amsfonts}
\usepackage{algorithmic}
\usepackage{graphicx}
\usepackage{textcomp}
\usepackage[dvipsnames,table]{xcolor}
\def\BibTeX{{\rm B\kern-.05em{\sc i\kern-.025em b}\kern-.08em T\kern-.1667em\lower.7ex\hbox{E}\kern-.125emX}}

\usepackage{microtype}
\usepackage{graphicx}
\usepackage{booktabs}
\usepackage{amsmath}
\usepackage{amssymb}
\usepackage{mathtools}
\usepackage{amsthm}
\usepackage{adjustbox}
\usepackage{color}
\usepackage{enumitem}
\usepackage{multirow}
\usepackage{makecell}
\usepackage{algorithmic}
\usepackage{algorithm}
\usepackage{etoolbox,siunitx}

\begin{document}

\title{MSC-Bench: Benchmarking and Analyzing Multi-Sensor Corruption for Driving Perception}

\author{Xiaoshuai Hao$^1$ \quad Guanqun Liu$^2$ \quad Yuting Zhao$^3$ \quad  Yuheng Ji$^3$ \quad  Mengchuan Wei$^4$ \quad Haimei Zhao$^5$ \\ Lingdong Kong$^6$ \quad Rong Yin$^7$ \quad Yu Liu$^8$ \\ Beijing Academy of Artificial Intelligence$^1$ \quad IQIYI$^2$ \quad Institute of Automation, CAS$^3$ \quad  Samsung$^4$ \\ The University of Sydney$^5$ \quad National University of Singapore$^6$ \\ Institute of Information Engineering, CAS$^7$ \quad Hefei University of Technology$^8$ 
}

\maketitle

\begin{abstract}
Multi-sensor fusion models play a crucial role in autonomous driving perception, particularly in tasks like 3D object detection and HD map construction. These models provide essential and comprehensive static environmental information for autonomous driving systems. While camera-LiDAR fusion methods have shown promising results by integrating data from both modalities, they often depend on complete sensor inputs. This reliance can lead to low robustness and potential failures when sensors are corrupted or missing, raising significant safety concerns.
To tackle this challenge, we introduce the Multi-Sensor Corruption Benchmark (MSC-Bench), the first comprehensive benchmark aimed at evaluating the robustness of multi-sensor autonomous driving perception models against various sensor corruptions. Our benchmark includes 16 combinations of corruption types that disrupt both camera and LiDAR inputs, either individually or concurrently.
Extensive evaluations of six 3D object detection models and four HD map construction models reveal substantial performance degradation under adverse weather conditions and sensor failures, underscoring critical safety issues. 
The benchmark toolkit and affiliated code and model checkpoints have been made publicly accessible\footnote[1]{https://msc-bench.github.io/}.
\end{abstract}

\begin{IEEEkeywords}
Autonomous Driving, Perception Robustness, 3D Object Detection, HD Map Construction, Multi-Sensor Corruption
\end{IEEEkeywords}
\section{Introduction}
\label{sec:intro}

The perception system is a critical component of autonomous vehicles, serving as the foundation for interaction between the vehicle and its driving environment. The system's performance—encompassing both accuracy and robustness—fundamentally influences the decision-making processes of autonomous vehicles. Notably, the robustness of perception algorithms is essential for the practical deployment of these vehicles, directly impacting the safety of future transportation systems for the general public.

Recently, researchers have developed fusion-based perception methods that integrate outputs from multiple sensors to enhance overall capabilities, leading to significant performance improvements across various tasks. For example, multi-sensor fusion approaches for 3D object detection and HD map construction have demonstrated superior accuracy compared to single-sensor methods that rely solely on cameras or LiDAR. However, these performance evaluations are typically conducted on clean datasets without any corruption, creating a gap in our understanding of the robustness of fusion-based perception methods under adverse conditions.

\begin{figure}[t]
    \centering
    \includegraphics[width=1.0\linewidth]{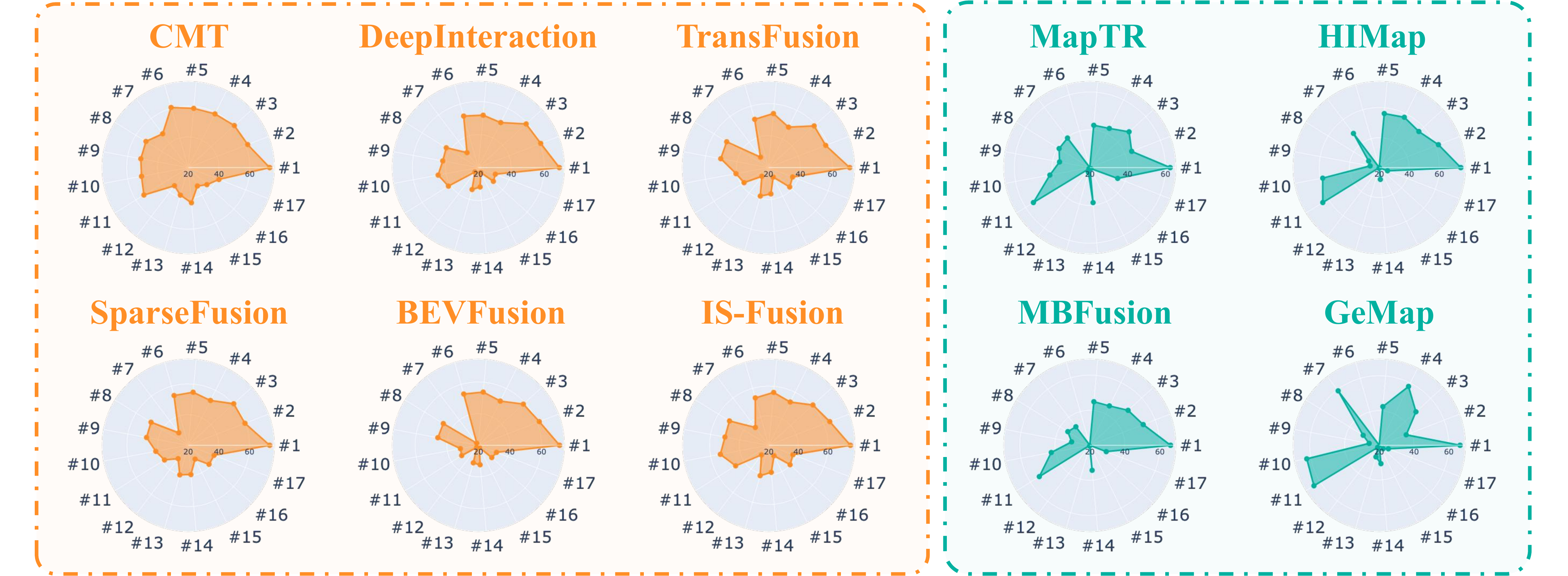}
    \caption{
    Radar charts display the performance of state-of-the-art multi-sensor 3D object detection models (left) and HD map construction models (right) under the Multi-Sensor Corruption Benchmark (MSC-Bench). We present NDS scores for 3D object detection methods and mAP scores for map construction methods across each corruption type and severity level.
    \textbf{MSC-Bench:} \textcolor{BurntOrange}{\#1} \texttt{Clean}, \textcolor{BurntOrange}{\#2} \texttt{Motion Blur}, \textcolor{BurntOrange}{\#3} \texttt{Temporal Misalignment}, \textcolor{BurntOrange}{\#4} \texttt{Spatial Misalignment}, \textcolor{BurntOrange}{\#5} \texttt{Fog}, \textcolor{BurntOrange}{\#6} \texttt{Snow}, \textcolor{BurntOrange}{\#7} \texttt{Camera Crash}, \textcolor{BurntOrange}{\#8} \texttt{Frame Lost}, \textcolor{BurntOrange}{\#9} \texttt{Cross Sensor}, \textcolor{BurntOrange}{\#10} \texttt{Cross Talk}, \textcolor{BurntOrange}{\#11} \texttt{Incomplete Echo}, \textcolor{BurntOrange}{\#12} \texttt{Camera Crash $\&$ Cross Sensor}, \textcolor{BurntOrange}{\#13} \texttt{Camera Crash $\&$ Cross Talk}, \textcolor{BurntOrange}{\#14} \texttt{Camera Crash $\&$ Incomplete Echo}, \textcolor{BurntOrange}{\#15} \texttt{Frame Lost $\&$  Cross Sensor}, \textcolor{BurntOrange}{\#16} \texttt{Frame Lost $\&$ Cross Talk} and \textcolor{BurntOrange}{\#17} \texttt{Frame Lost $\&$ Incomplete Echo}. The radius of each chart is normalized based on the \texttt{Clean} score. The larger the area coverage, the better the overall robustness.
}
    \label{motivation}
    \vspace{-1.5em}
\end{figure}

\begin{figure*}[t]
    \centering
    \includegraphics[width=0.97\textwidth]{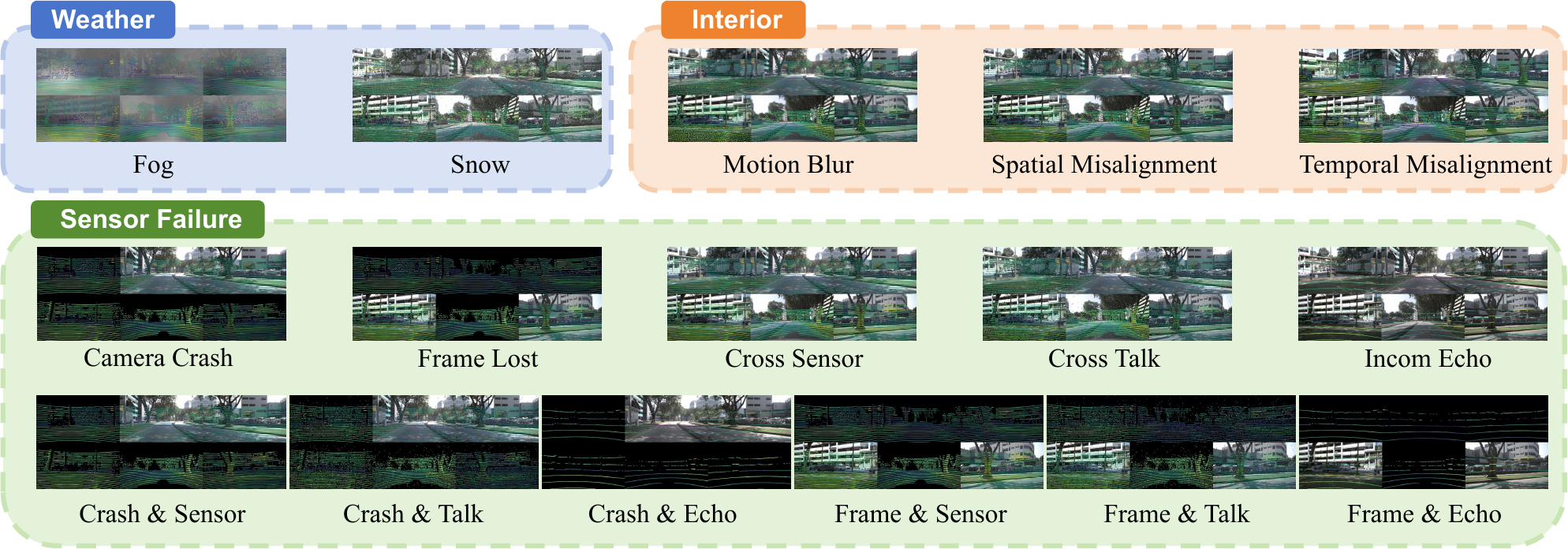}
    \caption{\textbf{Overview of the MSC-Bench.} Definitions of the multi-sensor corruptions in MSC-Bench. Our benchmark encompasses a total of 16 corruption types for multi-modal perception models, which can be categorized into weather, interior, and sensor failure scenarios. 
    }
    \label{fig2}
\vspace{-1em}
\end{figure*}

The robustness of perception algorithms refers to their performance in adverse conditions, including challenging driving environments, complex scenarios, and sensor failures. Unlike single-sensor algorithms, multi-sensor perception systems face a broader range of issues, such as misalignment and synchronization problems. Additionally, adverse conditions like fog or snow can affect sensors differently. Understanding how these factors impact multi-sensor performance and whether sensor fusion can mitigate these effects is essential and requires thorough investigation.

In this paper, we introduce 16 types of corruption specific to multi-sensor perception algorithms and evaluate the robustness of fusion-based methods across two autonomous driving tasks: six 3D object detection methods and four HD map construction methods. Results, as shown in Fig.~\ref{motivation}, reveal significant performance discrepancies between "clean" and corrupted datasets. Key findings include: 1) Camera-LiDAR fusion methods achieve strong performance by leveraging complementary information but often rely on complete sensor data, making them vulnerable to disruptions. 2) In 3D object detection, sensor failures and misalignments degrade performance, particularly under simultaneous disruptions like Frame Lost $\&$ Cross Sensor and Camera Crash $\&$ Cross Sensor, indicating a lack of adequate domain transfer and generalization. 3) In HD map construction, adverse weather, especially snow corruption, poses the greatest challenge by obscuring roads and reducing LiDAR reflectance, while Frame Lost $\&$ Cross Sensor conditions further emphasize the detrimental effects of dual-source information loss. In summary, both 3D object detection and HD map construction models are highly susceptible to sensor corruption, particularly from dual-source disruptions; future fusion models should enhance robustness against LiDAR variations and adapt to partial or missing camera data to improve reliability in real-world scenarios.
Through extensive benchmark studies, we further unveil crucial factors for enhancing the reliability of multi-sensor perception models against sensor corruption. The key contributions of this work are three-fold:

\begin{itemize}
    \item We introduce \textit{Multi-Sensor Corruption Benchmark (MSC-Bench)}, making the first attempt to comprehensively benchmark and evaluate the robustness of multi-sensor autonomous driving perception models against various sensor corruptions.

    \item 
    We analyze six 3D object detection models and four HD map construction models using MSC-Bench, offering valuable insights into design choices that enhance the robustness of multi-modal models.

    \item 
    We will provide our data generation source code and benchmark, allowing for the reproducibility of the results presented in this study, which will serve as a valuable contribution to the field.

\end{itemize}
\section{Related Work}

\begin{table*}[t]
    \caption{\textbf{Corruption Methods Overview}: Types, modalities, descriptions, and configurations of three severity levels of corruption.}
    \label{table:corruption_methods}
    \centering
    \begin{tabular}{
            p{0.16\linewidth}
            p{0.05\linewidth}
            p{0.40\linewidth}
            p{0.06\linewidth}
            p{0.06\linewidth}
            p{0.06\linewidth}}
        \toprule
        \textbf{Corruption} & \textbf{Modality} & \textbf{Description} & {\textbf{Level 1}} & {\textbf{Level 2}} & {\textbf{Level 3}} \\
        \midrule
        Camera Crash & C & Dropping view images & 2 & 4 & 5 \\
        Frame Lost & C & Dropping temporal frames & 2/6 & 4/6 & 5/6 \\
        Cross Sensor & L & Cross-sensor data by the number of beams to drop & 8 & 16 & 20 \\
        Crosstalk & L & Light impulse interference by adjusting the percentag & 0.03 & 0.07 & 0.12 \\
        Incomplete Echo & L & Incomplete LiDAR readings by adjusting the drop ratio & 0.75 & 0.85 & 0.95 \\
        Temporal Misalignment & LC & Frozen frame applied with probability $p$ & 0.2 & 0.4 & 0.6 \\
        Spatial Misalignment & LC & Extrinsic misalignment in degrees applied with probability $p$ & 1°, 0.2 & 2°, 0.4 & 3°, 0.6 \\
        Motion Blur & LC & Jitter noise from a Gaussian distribution with $\sigma_t$ & 0.06 & 0.10 & 0.13 \\
        Fog & LC & Approximated visibility in meters & 300 m & 150 m & 50 m \\
        Snow & LC & Approximated snowfall intensity in mm/h & 5 mm/h & 35 mm/h & 70 mm/h \\
        \bottomrule
    \end{tabular}
\label{tab1}
\end{table*}%

\noindent\textbf{Multi-Sensor 3D Object Detection}
The 3D object detection task focuses on identifying and localizing objects in three-dimensional space by predicting their 3D bounding boxes and categories using data from sensors like LiDAR and cameras, which is crucial for applications such as autonomous driving and robotics. While early methods relied on single sensors, the release of extensive autonomous driving datasets has spurred research into multi-sensor fusion for enhanced accuracy. Recent approaches include BEVFusion \cite{liu2023bevfusion}, which extracts features from both cameras and LiDAR using a Bird's-Eye View (BEV) space; DeepInteraction \cite{yang2022deepinteraction}, which facilitates interactions between modality-specific representations; and TransFusion \cite{22cvprtransfusion}, which uses a transformer-based mechanism for adaptive fusion. Other methods, such as SparseFusion \cite{xie2023sparsefusion}, utilize parallel detectors for instance-level fusion, while CMT \cite{yan2023crosscmt} incorporates a Coordinates Encoding Module for position-aware features. Is-Fusion \cite{isfusion} further improves detection by integrating scene-level and instance-level fusion to enhance feature collaboration.

\noindent\textbf{Multi-Sensor HD Map Construction}
The HD map construction task involves creating high-resolution maps that provide detailed vectorized representations of geometric and semantic information, such as lane boundaries and road structures, which are essential for accurate localization and path planning in autonomous driving. Recent camera-LiDAR fusion methods \cite{li2022hdmapnet,hao2025mapdistill,liu2023vectormapnet,MapTR} leverage the semantic richness of camera data and the geometric precision of LiDAR. BEV-level fusion, which combines inputs from both sensors into a shared BEV space, has gained attention \cite{liu2023bevfusion} for effectively integrating complementary features. However, existing methods often depend on complete sensor data, making them less robust to missing or corrupted information, which can lead to significant performance degradation. This paper focuses on  evaluating the robustness of multi-modal HD map construction.

\noindent\textbf{Driving Perception Robustness}
Researchers have recently focused on the robustness of various autonomous driving perception tasks. Studies like RoBoBEV \cite{xie2023robobev} evaluate the robustness of BEV perception tasks, while others aim to develop more resilient models and strategies. Robo3D \cite{kong2023robo3d} benchmarks LiDAR-based semantic segmentation and 3D object detection under sensor failures. Zhu et al. \cite{Zhu_2023_CVPR} assess the natural and adversarial robustness of BEV models, introducing a 3D consistent patch attack for spatiotemporal realism. PointDR \cite{xiao20233d} and UniMix \cite{Zhao_2024_CVPR} propose domain-adaptive methods for enhancing 3D semantic segmentation in adverse conditions. MapBench \cite{hao2024your} and Multi-corrupt \cite{multi-corrupt} offer benchmarks for evaluating the robustness of HD map construction and 3D object detection, respectively. In contrast to previous work, we present a more comprehensive benchmark that incorporates multi-sensor corruptions for fusion-based autonomous driving perception models, covering both HD map construction and 3D object detection tasks.
\section{Benchmarking Multi-Sensor Corruption}

\subsection{Multi-Sensor Corruption Definition}
The Multi-Sensor Corruption Benchmark (MSC-Bench) includes 16 corruption types, categorized into weather, interior, and sensor failure scenarios (see Fig. \ref{fig2}). It is constructed by corrupting the \textit{val} set of nuScenes\cite{20cvprnuscense}. Definitions of the corruption types can be found in Tab.~\ref{tab1}, with additional details provided below.

\begin{itemize}
   \item \textbf{Camera Crash}: Simulates continuous loss of images from certain viewpoints due to camera failure. Determine the level of corruption based on the number of dropped cameras. Note that this type of corruption applies only to camera sensors, while the LiDAR sensor remains clean.
    
    \item \textbf{Frame Lost}:
    Represents random frame loss to assess the model’s resilience to intermittent data loss, with the corruption level determined by the probability of frame dropping. 
    Note that this type of corruption applies only to camera sensors, while the LiDAR sensor remains clean.

    \item \textbf{Cross Sensor}: Arises due to the large variety of LiDAR sensor configurations (e.g., beam
    number, field-of-view, and sampling frequency).
    Determine the level of corruption based on the number of beams dropped.
    Note that this type of corruption applies only to LiDAR sensors, while the camera sensor remains clean.
    
    \item \textbf{Crosstalk}: Creates noisy points within the mid-range areas between two (or multiple)
    sensors, simulating interference.
    Determine the level of corruption by adjusting the percentage of light impulse interference.
    Note that this type of corruption applies only to LiDAR sensors, while the camera sensor remains clean.

    \item \textbf{Incomplete Echo}: 
    Represents incomplete LiDAR readings in some scan echoes. The level of corruption is determined by adjusting the drop ratio of these readings.
    Note that this type of corruption applies only to LiDAR sensors, while the camera sensor remains clean.
    
    \item \textbf{Fog}: 
    We use a fog simulator~\cite{hahner2021fog} to simulate LiDAR fog corruption. To maintain scene consistency between images and point clouds, we adapt the LiDAR fog parameters for the image fog generation process.
    
    \item  \textbf{Snow}: 
     We use a snow simulator \cite{hahner2022lidar} that models snow particles as opaque spheres and computes the reflection properties of wet surfaces, enabling us to corrupt the point cloud and image data based on snowfall levels.

    \item \textbf{Motion Blur}: 
    To replicate intense motion, vibrations, and the rolling shutter effect, we introduce jitter noise from a Gaussian distribution with a standard deviation of $\sigma_t$ into both point cloud and image data.

    \item \textbf{Spatial Misalignment}: 
    We introduce translation and rotation misalignment, creating a spatial offset between point cloud and camera inputs. We adjust the rotation angle and the proportion of affected data based on the severity level.

    \item \textbf{Temporal Misalignment}: 
    Timestamps from modalities like LiDAR and cameras are not always perfectly synchronized, so we introduce temporal misalignment to both the camera and point cloud data.

    \item \textbf{Camera Crash $\&$ Cross Sensor}: 
    For the camera sensor, we apply Camera Crash corruption, while for the LiDAR sensor, we use Cross Sensor corruption.

    \item \textbf{Camera Crash $\&$ Cross Talk}: 
   For the camera sensor, we use Camera Crash corruption, and for the LiDAR sensor, we apply Crosstalk corruption.

    \item \textbf{Camera Crash $\&$ Incomplete Echo}: 
    For the camera sensor, we apply Camera Crash corruption, and for the LiDAR sensor, we use Incomplete Echo corruption.

    \item \textbf{Frame Lost $\&$ Cross Sensor}: 
    For the camera sensor, we apply Frame Lost corruption, and for the LiDAR sensor, we use Cross Sensor corruption.
    
    \item \textbf{Frame Lost $\&$ Cross Talk}: 
    For the camera sensor, we use Frame Lost corruption, and for the LiDAR sensor, we use Cross Talk corruption.

    \item \textbf{Frame Lost $\&$ Incomplete Echo}: 
   For the camera sensor, we apply Frame Lost corruption, and for the LiDAR sensor, we use Incomplete Echo corruption.
\end{itemize}

\subsection{Robustness Evaluation Metrics}
To compare the robustness of different 3D object detectors and HD map constructors in multi-modal corrupted scenarios, we introduce two robustness evaluation metrics.

{\bf Resilience Score (RS)} 
We define $\mathrm{RS}$ as the relative robustness indicator for measuring how much accuracy a model can retain when evaluated on the corruption sets, which are calculated as follows:
\begin{equation}
    \mathrm{RS}_i=\frac{\sum_{l=1}^3\mathrm{Acc}_{i,l}}{3\times\mathrm{Acc}^\mathrm{clean}} ,\quad\mathrm{mRS}=\frac1N\sum_{i=1}^N\mathrm{RS}_i ,
\end{equation}
where $\mathrm{Acc}_{i,l}$ denotes the task-specific accuracy scores, with NDS (NuScenes Detection Score) for 3D object detection and mAP (mean Average Precision) for HD map construction, on corruption type $i$ at severity level $l$. $N$ is the total number of corruption types, and $\mathrm{Acc}^\mathrm{clean}$ denotes the accuracy score on the ``clean'' evaluation set. 
$\mathrm{mRS}$ (mean Resilience Score) represents the average score, providing an overall measure of the model's robustness across all types of corruption.

{\bf Relative Resilience Score (RRS)}
We define $\mathrm{RRS}$ as the critical metric for comparing the relative robustness of candidate models with the baseline model and $\mathrm{mRRS}$ as an overall metric to indicate the relative resilience score. The $\mathrm{RRS}$ and $\mathrm{mRRS}$ scores are calculated as follows:
\begin{equation}
    \mathrm{RRS}_i=\frac{\sum_{l=1}^3\mathrm{Acc}_{i,l}}{\sum_{l=1}^3\mathrm{Acc}_{i,l}^\mathrm{base}} -1,\quad\mathrm{mRRS}=\frac1N\sum_{i=1}^N\mathrm{RRS}_i ,
\end{equation}
where $\mathrm{Acc}_{i,l}^\mathrm{base}$ denotes the accuracy score of the baseline model.

\begin{table}[t]
\centering
\caption{
    \textbf{Benchmarking 3D object detection models.} 
We report detailed information on the methods grouped by $^1$ input modality, $^2$ backbone, and  $^3$ input image size. "L" and "C" represent LiDAR and camera, respectively. 
`Swin-T'', ``R50'', ``VoV-99'', and ``SEC'' are short for Swin-Transformer, ResNet50, VoVNet, and SECOND.  
We report nuScenes Detection Score (NDS) and mean Average Precision (mAP) on  the official nuScenes validation set.
}
\vspace{-0.1cm}
\scalebox{0.572}{
    \begin{tabular}{r|r|c|c|c|cc|cc}
    \toprule
    \rowcolor{gray!10}\textbf{Method} & \textbf{Venue} & \textbf{Modal}  & \textbf{Backbone} & \textbf{Image Size} & {\textbf{NDS}}$\uparrow$ &{\textbf{mAP}}$\uparrow$  & \textbf{mRS}$\uparrow$ & \textbf{mRRS}$\uparrow$
    \\\midrule\midrule
    BEVFusion~\cite{liu2023bevfusion} & ICRA'23 & {C \& L} & {Swin-T \& SEC} & $704 \times 256$ & $71.44$ & $68.72$  & $54.88$ & $0.00$
    \\
   SparseFusion~\cite{xie2023sparsefusion} & ICCV'23 & {C \& L} & {Swin-T \& SEC} & $800 \times 448$ & $73.15$ & $71.02$  & $60.11$ & $17.01$
    \\
   TransFusion~\cite{22cvprtransfusion} & CVPR'22 & {C \& L} & {R50 \& SEC} & {$800 \times 448$} & {$70.84$} &{$66.72$}  & {$60.12$} & {$12.30$}
     \\
   DeepInteraction~\cite{yang2022deepinteraction} & NIPS'22 & {C \& L} & {R50 \& SEC} & {$800 \times 448$}& {$69.09$} & {$68.72$} & {$59.01$} & {$6.93$}
     \\
  CMT~\cite{yan2023crosscmt} & ICCV'23 & {C \& L} & {VoV-99 \& SEC} & {$1600 \times 640$} & {$72.90$} &{$70.30$}  & {$67.17$} & {$32.93$}
     \\
   Is-Fusion~\cite{isfusion} & CVPR'24 & {C \& L}& {Swin-T \& SEC} & {$1056 \times 384$} & {$74.00$}  & {$72.8$} & {$62.10$} & {$22.42$}
    \\\bottomrule
    \end{tabular}}
    \label{tab2}
\end{table}
\begin{table}[t]
\centering
\caption{
    \textbf{Benchmarking HD map constructors.} 
    We report detailed information on the methods grouped by $^1$ input modality, $^2$ BEV encoder, $^3$ backbone, and $^4$ training epochs. "L" and "C" represent LiDAR and camera, respectively. "Effi-B0," "R50," "PP," and "SEC" refer to EfficientNet-B0, ResNet50, PointPillars, and SECOND. \texttt{AP} denotes performance on the clean nuScenes \textit{val} set. The subscripts $b.$, $p.$, and $d.$ denote \textit{boundary}, \textit{pedestrian crossing}, and \textit{divider}, respectively.
}
\vspace{-0.1cm}
\scalebox{0.572}{
    \begin{tabular}{r|r|c|c|c|cccc|cc}
    \toprule
    \rowcolor{gray!10}\textbf{Method} & \textbf{Venue} & \textbf{Modal}  & \textbf{Backbone} & \textbf{Epoch} & {\textbf{AP}$_{p.}$}$\uparrow$ &{\textbf{AP}$_{d.}$}$\uparrow$ & {\textbf{AP}$_{b.}$}$\uparrow$ & \textbf{mAP}$\uparrow$ & \textbf{mRS}$\uparrow$ & \textbf{mRSS}$\uparrow$
    \\\midrule\midrule
    \hline
    MapTR~\cite{MapTR} & ICLR'23 & {C \& L} & {R50 \& SEC} & {$24$} & {$55.9$} &{$62.3$} & {$69.3$} & {$62.5$} & {$55.91$} & {$0.00$}
     \\
    MBFusion~\cite{hao2024mbfusion} & ICRA'24 & {C \& L} & {R50 \& SEC} & {$24$} & {$61.6$} &{$64.4$} & {$72.5$} & {$66.1$} & {$50.83$} & {$-4.46$}
     \\
GeMap~\cite{GeMap} & ECCV'24 & {C \& L} & {R50 \& SEC} & {$24$} & {$66.3$} &{$62.2$} & {$71.1$} & {$66.5$} & {$55.25$} & {$4.77$}
     \\
     HIMap~\cite{HIMap} & CVPR'24 & {C \& L} & {R50 \& SEC} & {$24$} & {$71.0$} & {$72.4$} & {$79.4$} & {$74.3$} & {$50.29$} & {$4.27$}
    \\\bottomrule
    \end{tabular}}
    \label{tab3}
\end{table}
    \begin{table*}[h]
\caption{\textbf{Robustness benchmark of state-of-the-art multi-modal methods under multi-sensor corruptions.} 
For the 3D object detection task, we use NDS as the metric. Additionally, we use mAP as the metric for the HD map construction task.
}
\label{tab:rra_nds}
\begin{adjustbox}{width=\textwidth}
\begin{tabular}{>{\centering\arraybackslash}p{1.5cm}|c|cccccccccccccccc|c}
\toprule
\multicolumn{2}{c|}{\textbf{Model}} & \begin{tabular}[c]{@{}c@{}}\textbf{Motion} \\ \textbf{Blur}\end{tabular} & \begin{tabular}[c]{@{}c@{}}\textbf{Temporal} \\ \textbf{Mis.}\end{tabular} & \begin{tabular}[c]{@{}c@{}}\textbf{Spatial} \\ \textbf{Mis.}\end{tabular} & \textbf{Fog} & \textbf{Snow} & \textbf{Camera Crash}& \textbf{Frame Lost} &  \textbf{Cross Sensor} &  \textbf{Cross Talk} & \textbf{Incomplete Echo} & \begin{tabular}[c]{@{}c@{}}\textbf{Camera Crash,} \\ \textbf{Cross Sensor}\end{tabular} & \begin{tabular}[c]{@{}c@{}}\textbf{Camera Crash,} \\ \textbf{Cross Talk}\end{tabular} & \begin{tabular}[c]{@{}c@{}}\textbf{Camera Crash,} \\ \textbf{Incomplete Echo}\end{tabular} & \begin{tabular}[c]{@{}c@{}}\textbf{Frame Lost,} \\ \textbf{Cross Sensor}\end{tabular} & \begin{tabular}[c]{@{}c@{}}\textbf{Frame Lost,} \\ \textbf{Cross Talk}\end{tabular} & \begin{tabular}[c]{@{}c@{}}\textbf{Frame Lost,} \\ \textbf{Incomplete Echo}\end{tabular} & \begin{tabular}[c]{@{}c@{}}\textbf{mRS$\uparrow$}\end{tabular} \\
\midrule\midrule
\multirow{4}{*}{\parbox[c][1.9cm][c]{1.5cm}{\centering 3D \\ Object \\ Detection}} 
& CMT~\cite{yan2023crosscmt} & 84.25  & 83.05  & 80.91  & 80.44  & 83.15  & 71.69  & 70.35  & 65.21  & 69.73  & 73.84  & 47.71  & 52.97  & 58.93  & 45.78  & 49.98  & 56.78  & 67.17 \\
& DeepInteraction~\cite{yang2022deepinteraction} &  87.64  & 85.69  & 73.02  & 75.03  & 75.87  & 61.97  & 60.68  & 45.36  & 64.76  & 59.76  & 33.84  & 48.94  & 46.00  & 33.80  & 46.80  & 44.97  & 59.00 \\
& TransFusion~\cite{22cvprtransfusion} & 82.50  & 82.88  & 68.15  & 76.15  & 72.58  & 71.54  & 71.03  & 39.82  & 57.34  & 53.53  & 37.90  & 54.39  & 51.59  & 37.68  & 53.61  & 51.16  & 60.12 \\
& SparseFusion~\cite{xie2023sparsefusion} & 81.61  & 82.25  & 71.93  & 74.79  & 73.42  & 66.28  & 65.12  & 41.26  & 56.69  & 52.21  & 42.27  & 54.75  & 53.52  & 41.10  & 52.50  & 52.10  & 60.11 \\
& BEVFusion~\cite{liu2023bevfusion} &85.92  & 82.01  & 71.66  & 75.26  & 75.16  & 64.47  & 64.48  & 30.25  & 44.14  & 45.29  & 30.25  & 44.14  & 45.30  & 30.25  & 44.15  & 45.29  & 54.88\\
& Is-Fusion~\cite{isfusion} & 86.82  & 81.61  & 71.12  & 75.46  & 71.97  & 69.22  & 67.75  & 47.40  & 72.05  & 62.62  & 38.11  & 55.87  & 51.84  & 37.70  & 53.36  & 50.74  & 62.10\\
\midrule
\multirow{3}{*}{\parbox[c][1.2cm][c]{1.5cm}{\centering HD \\ Map \\ Construction}} 
& MapTR~\cite{MapTR} & 70.00 & 76.94 & 69.05 & 67.94 & 19.55 & 62.56 & 58.08 & 63.34& 66.40& 88.16& 33.28 & 36.32 & 61.76 & 30.56 & 33.28 & 57.28 & 55.91 \\
& HIMap~\cite{HIMap} & 83.77 & 74.93 & 77.31 & 75.56 & 23.79 & 38.09 & 35.26 & 65.28 & 78.47 & 86.41 & 19.25 & 27.59 & 37.55& 19.38 & 27.19& 34.86 & 50.29 \\
& MBFusion~\cite{hao2024mbfusion} & 79.69 & 74.96 & 68.19 & 67.97 & 23.57 & 52.60 & 46.25 & 50.26& 64.13 & 81.56 & 25.47 & 30.97 & 51.92 & 22.68 & 27.67 & 45.47 & 50.83 \\
& GeMap~\cite{GeMap} & 55.08 & 72.99& 86.87 & 63.63 & 19.58& 46.44 & 38.98 & 89.02  & 93.83 & 96.52 & 32.37 & 40.51 & 46.01 & 29.05 & 34.58 & 38.58& 55.25\\
\bottomrule
\end{tabular}
\end{adjustbox}
\vspace{-0.5em}
\end{table*}

\begin{figure*}[t]
    \centering
    \setlength{\abovecaptionskip}{-0.01cm}  
    \includegraphics[width=1.0\textwidth]{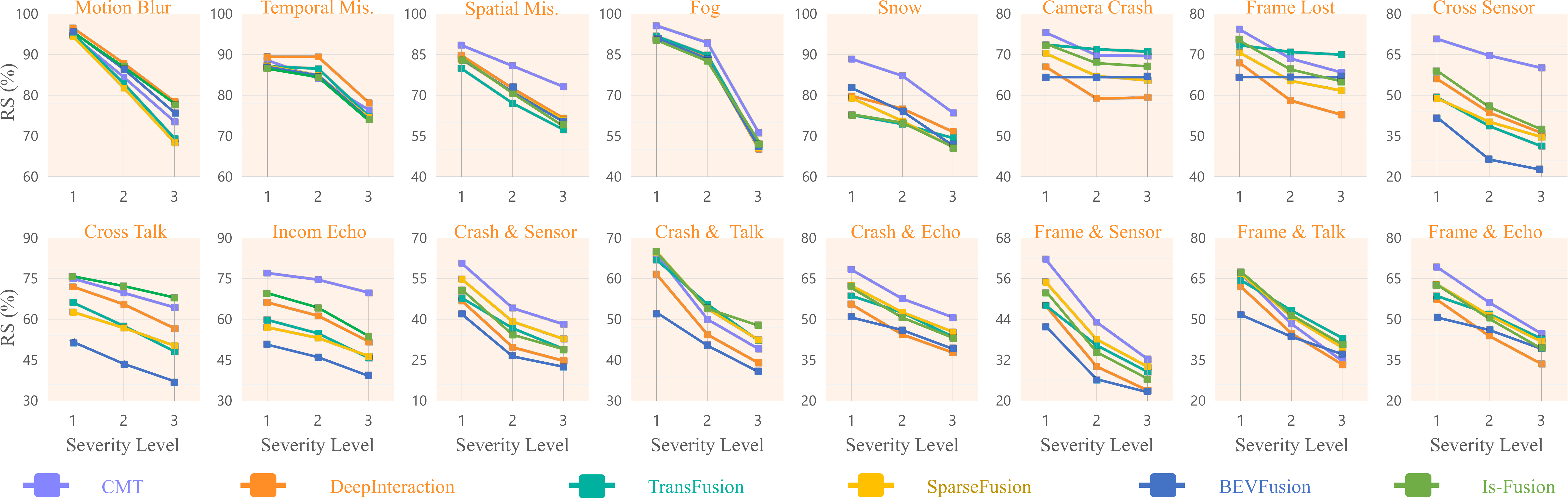}
    \caption{
    Robustness against all corruption types and severity levels in 3D object detection tasks is evaluated through the Resilience Score (RS), calculated using the NDS score for varying severity levels.
    }
    \label{fig3}
    \vspace{-1em}
\end{figure*}

\begin{figure*}[t]
    \centering
     \setlength{\abovecaptionskip}{-0.01cm}  
     \includegraphics[width=1.0\textwidth]{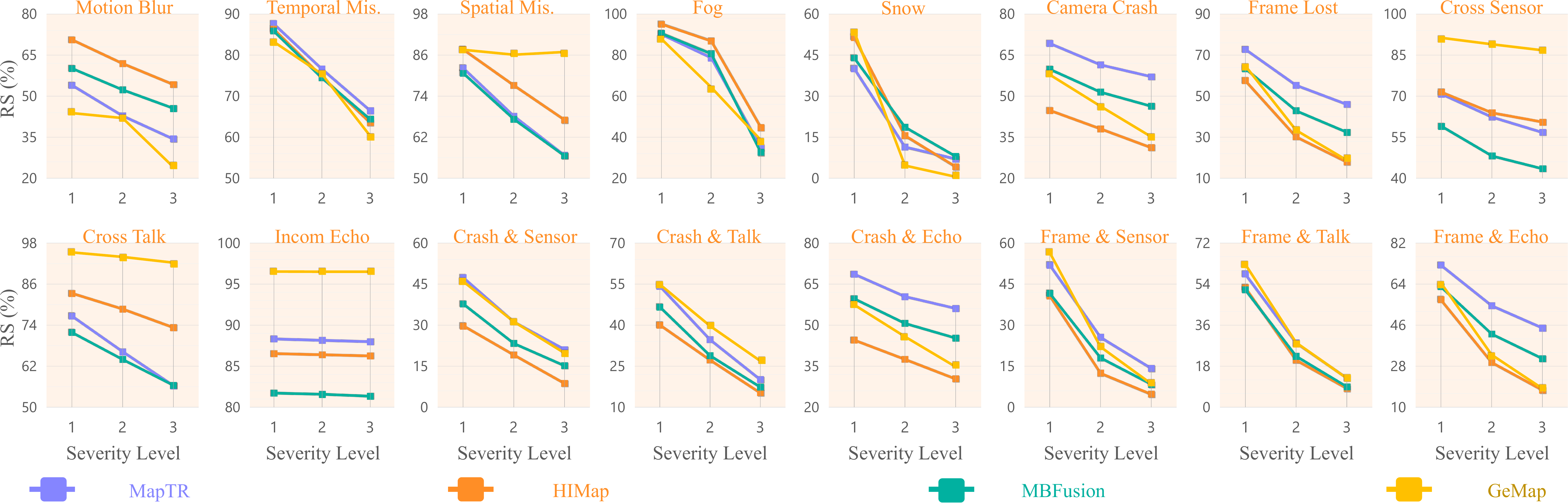}
    \caption{
    Robustness against all corruption types and severity levels in HD map construction tasks is assessed using the Resilience Score (RS), calculated based on the mAP score for different severity levels.
    }
    \vspace{-1.0em}
    \label{fig4}

\end{figure*}

\begin{figure*}[t]
    \centering
    \includegraphics[width=0.95\textwidth]{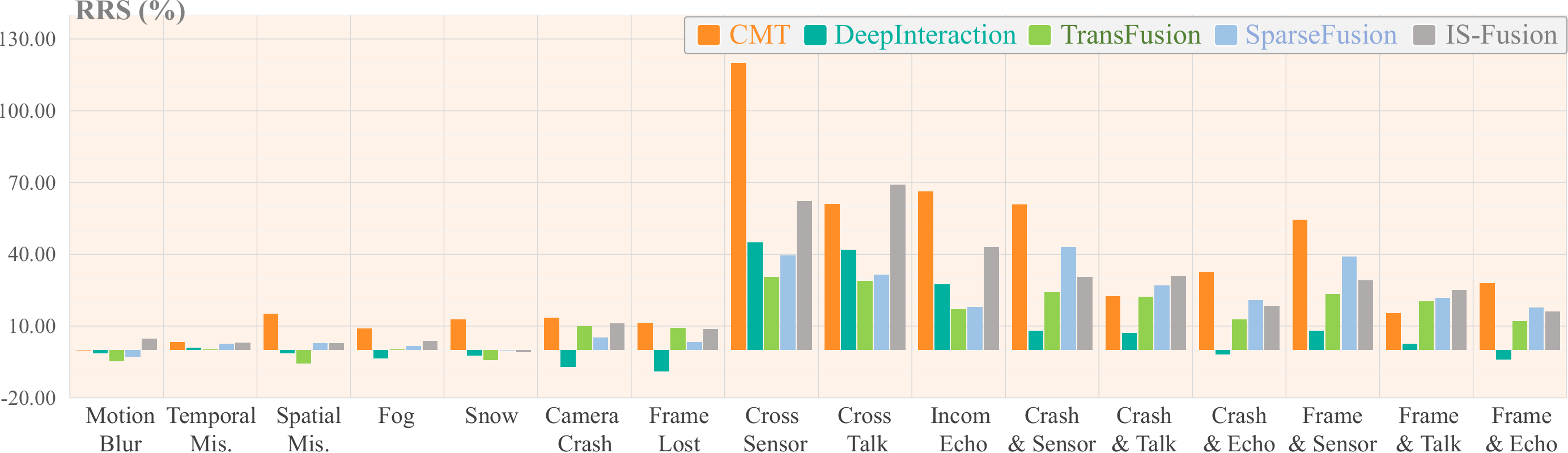}
    \caption{Relative robustness visualization. Relative Resilience Score (RRS) computed with NDS using BEVFusion~\cite{liu2023bevfusion} as baseline. 
    }
    \label{fig5}
     \vspace{-1em}
\end{figure*}

\begin{figure*}[t]
    \centering
    \includegraphics[width=0.95\textwidth]{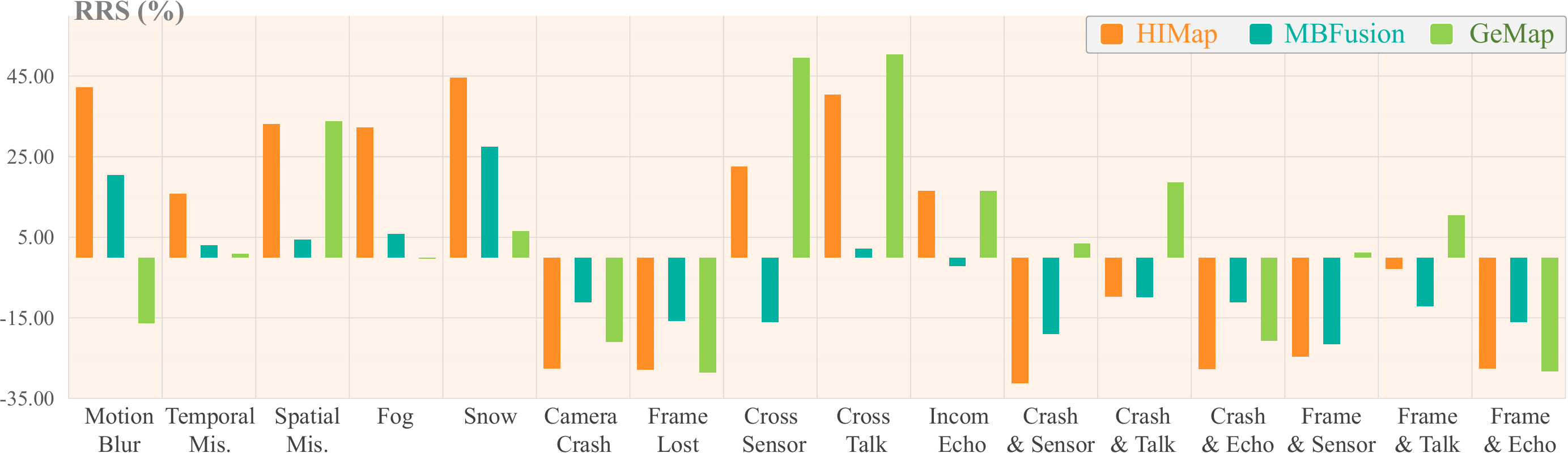}
    \caption{Relative robustness visualization. Relative Resilience Score (RRS) computed with mAP using MapTR~\cite{MapTR} as the baseline. 
    }
    \label{fig6}
   \vspace{-1.1em}
\end{figure*}

\section{Experiments and Analysis}
\subsection{Benchmarking Multi-Sensor 3D Object Detection}
\textbf{Candidate models}
Our MSC-Bench includes a total of six multi-sensor 3D object detection models: CMT~\cite{yan2023crosscmt}, DeepInteraction~\cite{yang2022deepinteraction}, TransFusion~\cite{22cvprtransfusion}, SparseFusion~\cite{xie2023sparsefusion},  BEVFusion~\cite{liu2023bevfusion} and Is-Fusion~\cite{isfusion}. We present the basic information for these models in Tab.~\ref{tab2}, including input modality, backbone, image size, and performance on the official nuScenes validation set.

\textbf{3D Object Detection Benchmarking Analysis}
We present the overall robustness benchmarking results, including $\mathrm{mRS}$ and $\mathrm{mRRS}$, for the six multi-sensor candidate models in Tab.~\ref{tab2}. The table shows that model robustness under corruption does not strongly correlate with performance on the clear validation set. For example, while Is-Fusion achieves the highest NDS and mAP scores, its $mRS$ and $mRRS$ scores are below expectations. In contrast, CMT exhibits excellent robustness, achieving the highest robustness scores.

To analyze the models' robustness across different corruption types, we present the Resilience Scores for 16 corruption types in Tab.~\ref{tab:rra_nds} (top) and illustrate robustness performance across varying severity levels in Fig.~\ref{fig3}. The data shows that sensor failure and misalignment-related corruptions, such as Cross Sensor, Camera Crash $\&$ Cross Sensor, and Frame Lost $\&$ Cross Sensor, significantly impact model performance. In contrast, individual sensor failures like Camera Crash, Frame Lost, and Incomplete Echo have minimal effects on robustness. However, when these failures occur simultaneously, as seen in Camera Crash $\&$ Incomplete Echo and Frame Lost $\&$ Incomplete Echo, model robustness is substantially compromised.

From Fig.~\ref{fig3}, most corruption types lead to a linear decline in model robustness as severity increases. However, the robustness degradation from Camera Crash and Frame Lost is relatively minor, showing a distinct pattern compared to other corruptions. Notably, for Temporal Misalignment and Fog, robustness remains stable at severity levels 1 and 2 but drops dramatically at level 3 as severity intensifies.

Fig.~\ref{fig5} shows the relative resilience scores of different models based on BEVFusion. DeepInteraction underperforms the base model in eight corruption types, while only CMT surpasses the base model across all corruption types, achieving the best performance in 12 of them.

\subsection{Benchmarking Multi-Sensor HD Map Construction}
\textbf{Candidate models}
Our MSC-Bench includes four multi-sensor HD map constructors: MapTR~\cite{MapTR}, HIMap~\cite{HIMap}, MBFusion~\cite{hao2024mbfusion}, and GeMap~\cite{GeMap}. We present the basic information for these models in Tab.~\ref{tab3}, including input modality, backbone, training epochs, and performance on the official nuScenes validation set.

\textbf{HD map construction Benchmarking Analysis}
Tab.~\ref{tab3} presents the overall robustness performance of the four multi-sensor HD map construction models, measured by $\mathrm{mRS}$ and $\mathrm{mRRS}$. MapTR and GeMap achieve comparable scores, outperforming MBFusion and HIMap. Performance on specific corruption types is detailed in Tab.~\ref{tab:rra_nds} (bottom), highlighting that Snow severely impacts all models, reducing mAP to a range of 19 to 24. Additionally, combinations of sensor corruptions, such as Camera Crash $\&$ Cross Sensor, Camera Crash $\&$ Cross Talk, Frame Lost $\&$ Cross Sensor, and Frame Lost $\&$ Cross Talk, lead to significant performance degradation.

Fig.~\ref{fig4} illustrates how the performance of the four models changes as corruption severity increases, with most models showing a linear decline. Notably, variations in the severity of Incomplete Echo have a negligible impact on all four models. Among the models, GeMap and MapTR achieve the best results in eight corruption types, demonstrating very similar performance on the aforementioned combination corruptions.
Fig.~\ref{fig6} shows the relative resilience scores of other HD map construction models compared to MapTR. GeMap, HIMap, and MBFusion underperform the base model in 5, 8, and 10 types of corruption, respectively.
This disparity highlights the varying levels of robustness among these models, emphasizing the need for effective strategies to enhance resilience against diverse types of sensor corruption.

\section{Conclusion}
In this paper, we introduced the Multi-Sensor Corruption Benchmark (MSC-Bench) to assess the robustness of multi-sensor autonomous driving perception models under 16 types of corruption. Our analysis of six 3D object detection models and four HD map construction models revealed significant performance discrepancies between clean and corrupted datasets, highlighting vulnerabilities to sensor disruptions, particularly dual-source failures like Frame Lost $\&$ Cross Sensor. While camera-LiDAR fusion methods demonstrated strong performance, they struggled with incomplete sensor data. Additionally, adverse weather conditions, especially snow, severely impacted HD map construction by obscuring critical elements. These findings underscore the need for more resilient fusion models that can effectively handle partial or missing sensor data and misalignment.

\bibliographystyle{IEEEbib}
\bibliography{icme2025references}

\end{document}